\def\confName{CVPR}
\def\confYear{2024}

\def\paperTitle{Plan, Posture and Go: Towards Open-World Text-to-Motion Generation}

\def\authorBlock{
    Jinpeng Liu$^{1}$\footnotemark[1] \qquad
    Wenxun Dai$^{1}$\footnotemark[1] \qquad
    Chunyu Wang$^{2}$\footnotemark[1] \qquad
    Yiji Cheng$^{1}$\qquad \\
    Yansong Tang$^{1}\footnotemark[2]$\qquad
    Xin Tong$^{2}$\qquad \\ \\
    $^{1}$Shenzhen International Graudate School, Tsinghua University \qquad $^{2}$Microsoft Research Asia \\ 
    {\tt\small \{liujp22@mails., daiwx23@mails., cyj22@mails., tang.yansong@sz.\}tsinghua.edu.cn} \\
    {\tt\small\{chnuwa, xtong\}@microsoft.com}
}

\newif\ifreview 
\newif\ifarxiv \newcommand{\arxiv}{\arxivtrue}
\newif\ifcamera 
\newif\ifrebuttal 

\arxiv
\pdfoutput=1
\documentclass[10pt,twocolumn,letterpaper]{article}
\ifreview \usepackage[review]{cvpr} \fi
\ifarxiv \usepackage[pagenumbers]{cvpr} \fi
\ifrebuttal \usepackage[rebuttal]{cvpr} \fi
\ifcamera \usepackage{cvpr} \fi

\usepackage{enumerate}
\usepackage{graphicx}	
\usepackage{amsmath}	
\usepackage{amssymb}	
\usepackage{booktabs}
\usepackage{times}
\usepackage{microtype}
\usepackage{epsfig}
\usepackage[table,xcdraw,dvipsnames]{xcolor}
\usepackage{caption}
\usepackage{float}
\usepackage{placeins}
\usepackage{color, colortbl}
\usepackage{stfloats}
\usepackage{enumitem}
\usepackage{tabularx}
\usepackage{xstring}
\usepackage{multirow}
\usepackage{xspace}
\usepackage{url}
\usepackage{subcaption}
\usepackage{xcolor}
\usepackage{listings}
\usepackage{cancel}
\usepackage{soul}
\usepackage{ulem}
\usepackage[hang,flushmargin]{footmisc}

\ifcamera \usepackage[accsupp]{axessibility} \fi

% ...

\ifarxiv  \fi

\newcommand{\R}[1]{{%
    \textbf{%
        \ifstrequal{#1}{1}{\textcolor{red}{R#1}}{%
        \ifstrequal{#1}{2}{\textcolor{blue}{R#1}}{%
        \ifstrequal{#1}{3}{\textcolor{magenta}{R#1}}{%
        \ifstrequal{#1}{4}{\textcolor{teal}{R#1}}{%
                           \textcolor{cyan}{R#1}%
        }}}}%
    }%
}}

\usepackage{xr-hyper}

\makeatletter
\newcommand*{\addFileDependency}[1]{
  \typeout{(#1)}
  \@addtofilelist{#1}
  \IfFileExists{#1}{}{\typeout{No file #1.}}
}

\makeatother

\definecolor{cvprblue}{rgb}{0.21,0.49,0.74}
\usepackage[pagebackref,breaklinks,colorlinks,citecolor=cvprblue]{hyperref}
\usepackage[capitalize]{cleveref}
\crefname{section}{Sec.}{Secs.}
\crefname{table}{Table}{Tables}
\crefname{figure}{Fig.}{Figs.}

\begin{document}
\title{\paperTitle}
\author{\authorBlock}
\twocolumn[{
    \renewcommand\twocolumn[1][]{#1}
    \maketitle
    \begin{center}
        \includegraphics[width=1.0\linewidth]{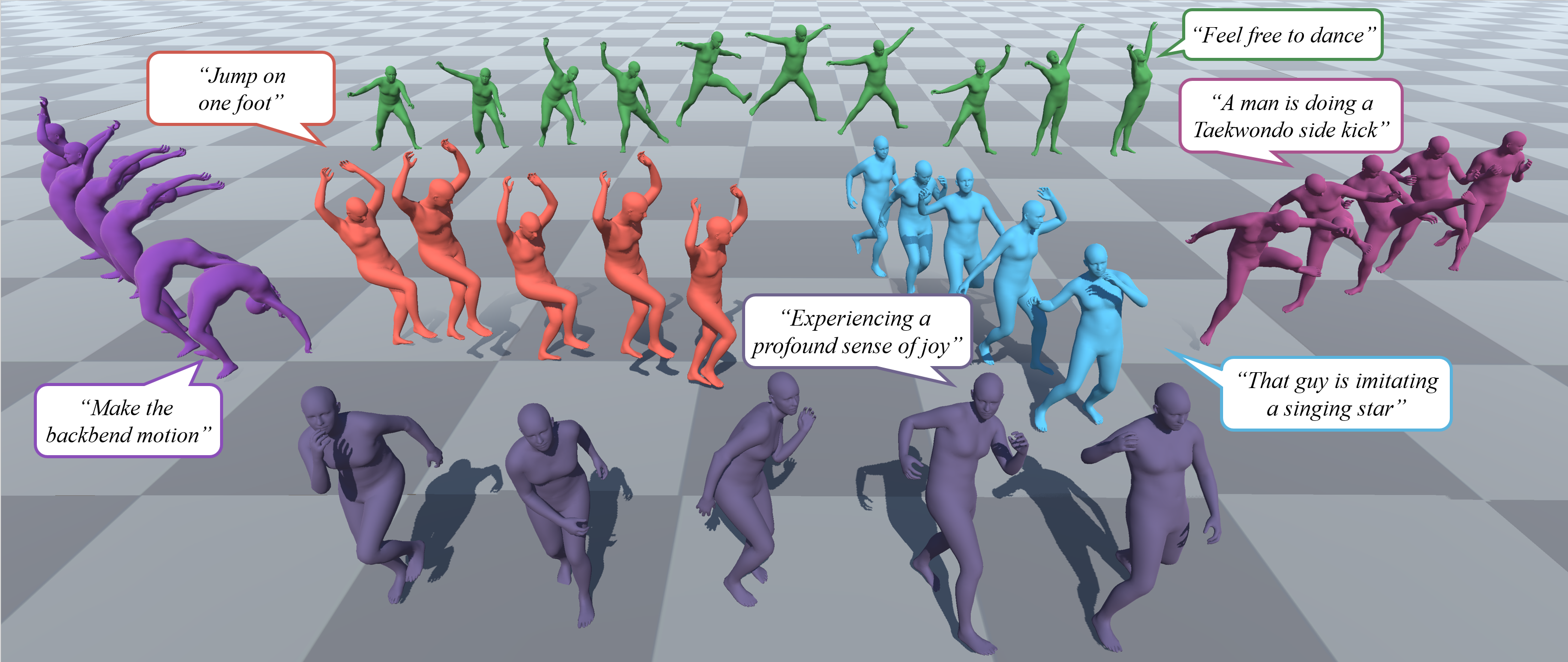}
        \captionof{figure}{
        Exemplary motions generated by our proposed PRO-Motion system.
        Different from conventional models trained on paired text-motion data, our PRO-Motion can generate 3D human motion with global body translation and rotation from \textit{open-world} text prompts, such as \textit{``Jump on one foot''} and \textit{``Experiencing a profound sense of joy''}.        
%
}
        \label{fig:teaser}
    \end{center}
}]
\renewcommand{\thefootnote}{\fnsymbol{footnote}}
\footnotetext[1]{Equal Contribution}\mbox{}\hfill \footnotetext[2]{Corresponding Author} 

\renewcommand*{\thefootnote}{\arabic{footnote}}
\begin{abstract}
Conventional text-to-motion generation methods are usually trained on limited text-motion pairs, making them hard to generalize to open-world scenarios. 
Some works use the CLIP model to align the motion space and the text space, aiming to enable motion generation from natural language motion descriptions. However, they are still constrained to generate limited and unrealistic in-place motions. 
To address these issues, we present a divide-and-conquer framework named \textbf{PRO-Motion}\footnote{\textbf{PRO-Motion}: \textbf{P}lan, postu\textbf{R}e and g\textbf{O} for text-to-\textbf{Motion} generation}, 
which consists of three modules as motion planner, posture-diffuser and go-diffuser.
The motion planner instructs Large Language Models (LLMs) to generate a sequence of scripts describing the key postures in the target motion. Differing from natural languages, the scripts can describe all possible postures following very simple text templates. This significantly reduces the complexity of posture-diffuser, which transforms a script to a posture, paving the way for open-world generation.
Finally,  go-diffuser, implemented as another diffusion model,  estimates whole-body translations and rotations for all postures, resulting in realistic motions. 
Experimental results have shown the superiority of our method with other counterparts, and demonstrated its capability of generating diverse and realistic motions from complex open-world prompts such as ``Experiencing a profound sense of joy''. The project page is available at \href{https://moonsliu.github.io/Pro-Motion/}{https://moonsliu.github.io/Pro-Motion/}.

\end{abstract}

\section{Introduction}
\label{sec:intro}

\begin{figure}[tbp]
    \centering
    \includegraphics[width=0.45\textwidth]{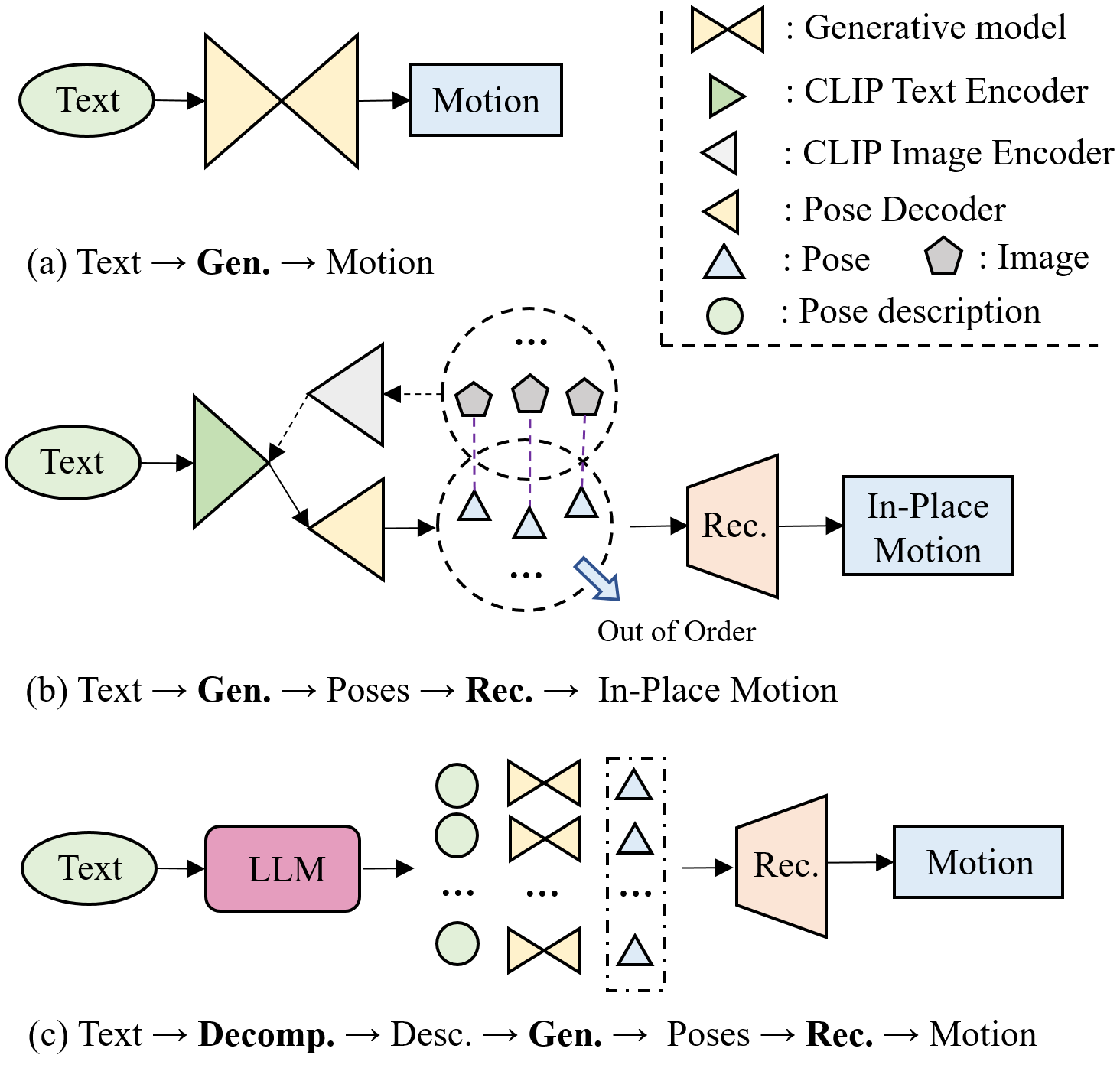}
    \caption{\textbf{Comparison of different paradigms for text-to-motion generation.} (a) Most existing models leverage the generative models~\cite{gan, vae, ddpm1} to construct the relationship between text and motion based on text-motion pairs. (b) Some methods render 3D poses to images and employ the image space of CLIP to align text with poses. Then they reconstruct the motion in the local dimension based on the poses. (c) Conversely, we decompose motion descriptions into structured pose descriptions. Then we generate poses based on corresponding pose descriptions. Finally, we reconstruct the motion in local and global dimensions. ``Gen.'', ``Decomp.'', ``Desc.'', ``Rec.'' stand for ``Generative model'', ``Decompose'', ``Pose Description'' and ``Reconstruction'' respectively.}
    \label{fig:formulation}
\end{figure}

Text-to-motion generation has attracted rapidly increasing attention~\cite{text2action, temos, mdm} due to its important roles in many applications such as virtual reality, video games, and the film industry. The prior models usually train GANs~\cite{text2action, dvgans}, VAEs~\cite{action2motion, actor, temos, teach} and Diffusion Models~\cite{mld, mdm, motiondiffuse, mofusion, diffprior, physdiff, remodiffuse} from paired text-motion data and have achieved reasonable generation results when the text prompts are similar as those in the training set. ~\cref{fig:formulation} (a) illustrates this paradigm. However, they struggle to handle open-world text prompts beyond the existing datasets, which is a core challenge that has to be addressed. Otherwise, they can only generate limited ``toy-like'' motions.

Some recent work~\cite{motionclip, avatarclip, oohmg} propose to enhance their model's ability to handle natural language motion descriptions beyond the training data. To that end, they leverage the pre-trained vision-language model CLIP~\cite{clip} to align the poses in the training motions with the motion descriptions, hoping to generate poses from natural languages. This is depicted in \cref{fig:formulation} (b). However, the text space of CLIP, which is learned on natural languages, is largely different from motion descriptions, making it ineffective to connect natural languages and motions. As a result, these methods are still constrained to generate motions from limited text prompts. Besides, due to the lack of temporal priors in CLIP, these methods have difficulty in generating motions with correct chronological order. As a result, they can only generate unrealistic in-place motions.

In this paper, we present a divide-and-conquer framework named \textbf{PRO-Motion}, which consists of three steps as \textbf{P}lan, postu\textbf{R}e, and \textbf{G}o for open-world text-to-\textbf{Motion} generation, as shown in \cref{fig:formulation} (c).
In the first ``plan'' stage, we introduce a motion planner that translates complex natural language motion descriptions into a sequence of posture scripts that describe body part relationships following a simple template, such as ``The man is standing upright, his torso is vertical. His left foot is slightly above the ground. His arms are relaxed at his sides''. This is realized by leveraging the motion commonsense in LLMs which is further enhanced by in-context demonstrations. It is important to understand that although the scripts are simple and limited to a small space, they are expressive to cover all possible postures due to their compositional nature. The motion planner bridges the gap between natural languages and pose descriptions and effectively addresses out-of-distribution problems.

%
%
%


Benefiting from the merits above, during the second ``posture'' stage, we can train a generative model to achieve script-to-posture generation only using a relatively small labeled dataset. We conjecture and demonstrate that the model has strong generalization capability and can cover extensive postures and scripts considering that a novel posture or script can be decomposed into multiple familiar body parts. In the implementation, we developed a diffusion-based model called posture-diffuser on a public dataset~\cite{posescript}, which perceives the connection between structured pose descriptions and body parts leading to diverse and realistic postures. Then we further utilize a posture planning module to select key poses, taking into account the consistency of adjacent poses and the semantic alignment between text and poses.

Furthermore, in the last ``go'' stage, we have observed that we can predict both translation and rotation by analyzing multiple consecutive body postures. For example, in a sequence where the initial pose depicts a standing pose followed by a left-foot step in the second pose and a right-foot step in the third pose, we can estimate forward translation. Additionally, learning interpolation between adjacent key poses is straightforward and requires only a small amount of motion data to capture such priors effectively. Accordingly, a transformer-based~\cite{attention} go-diffuser module is designed to capture the inner connection between key poses.

To verify the effectiveness of our proposed PRO-Motion, we conduct experiments on a variety of datasets. Both quantitative and qualitative results have shown the advantage of our method compared with the state-of-the-art approaches for open-world text-to-motion generation and demonstrated its capability of generating diverse and realistic motions from complex prompts such as \textit{``Jump on one foot''} and \textit{``Experiencing a profound sense of joy''}.

\section{Related Work}
\label{sec:related}

\textbf{Text-to-Motion Generation.} Based on existing labeled motion capture datasets~\cite{kit, amass, babel, humanml3d, flag3d, action2motion, uestc, ntu, motionx}, existing works have explored various generative models for text-driven motion generation, such as GANs~\cite{text2action, dvgans}, VAEs~\cite{action2motion, actor, temos, teach, motionclip} and Diffusion Models~\cite{mld, mdm, motiondiffuse, mofusion, diffprior, physdiff, remodiffuse}. However, these methods are constrained by the heavy reliance on limited text-motion paired datasets.

To tackle this problem, some works~\cite{avatarclip, oohmg} try to leverage the current powerful large-scale pre-trained models, \ie, CLIP \cite{clip}, to overcome the data limitation and achieve open-vocabulary motion generation. AvatarCLIP~\cite{avatarclip} generates motions for given textual descriptions through online matching and optimization. Nevertheless, matching is unable to generate out-of-distribution candidate poses, which limits the ability to generate complex motions, and online optimization is time-consuming and unstable. OOHMG~\cite{oohmg} uses CLIP image features to generate candidate poses and performs motion generation via mask learning. However, this method cannot capture the chronological order of actions due to the lack of temporal priors in CLIP, leading to inaccurate or even completely opposite motion. Our approach takes a different step to probe the powerful prior knowledge of human body pose and motion in LLMs to enhance text-motion alignment capability and enable open-world motion generation. 

\noindent \textbf{Keyframe-based Motion Generation.} Given that motion can be viewed as a composition of a sequence of poses, keyframe-based motion generation has attracted lots of interest. Motion prediction involves generating unrestricted motion continuation when provided with one or more keyframes of animation as context.  Early efforts~\cite{motion_prediction_rnn1, motion_prediction_rnn2, motion_prediction_rnn3, motion_prediction_rnn4, motion_prediction_rnn5, motion_prediction_rnn6, motion_prediction_rnn7, motion_prediction_rnn8} employed RNNs to model human motion sequence, motivated by the powerful capability in capturing temporal dynamics. Besides RNNs, other network architectures like CNNs~\cite{motion_prediction_cnn1, motion_prediction_cnn2} and GCNs~\cite{motion_prediction_gcn1} are proposed to enhance the modeling of temporal and movement relationships. The emergence of Transformers~\cite{motion_prediction_ts1, motion_prediction_ts2} has further facilitated the modeling of long-range dependencies within a motion sequence. Close to our method is motion in-betweening, which is constrained by both past and future keyframes. Early methods include physically-based approaches~\cite{motion_inb_phc1, motion_inb_phc2, motion_inb_phc3} that involve solving optimization problems, as well as statistical models~\cite{motion_inb_st1, motion_inb_st2, motion_inb_st3}. More recently, some neural network-based methods such as RNNs~\cite{motion_inb_rnn1, motion_inb_rnn2, motion_inb_rnn3}, CNNs~\cite{motion_inb_cnn1, motion_inb_cnn2, motion_inb_cnn3}, and Transformers~\cite{motion_inb_ts1, motion_inb_st2, motion_inb_ts3} have gained dominance in this field. Unlike motion in-betweening methods that explicitly provide translation and rotation, we achieve the prediction of translation and rotation, as well as pose interpolation, by having the model learn priors between adjacent key poses.

\noindent \textbf{LLM aided Visual Content Generation.} In recent years, large language models (LLMs) \cite{gpt3, instructgpt, gpt4, opt, llama, glm} have attracted substantial interest in the field of natural language processing (NLP) and artificial general intelligence (AGI) owing to their remarkable proficiency in tasks such as language generation, reasoning, world knowledge, and in-context learning. \cite{instructpix2pix, promptist} combine large language models with diffusion-based generative models \cite{ddpm1, sd} aimed at generating prompts for higher-quality image generation. \cite{videodirectorgpt, direct2v} leverage large language models to plan the generation of visual content and identify the pivotal actions, enabling complex dynamic video generation. Another line of works, including \cite{visualchatgpt, mm-react, hugginggpt, internchat, videochat}, have proposed to integrate visual APIs with language models to facilitate decision-making or planning based on visual information, which further connects vision and language models. Close to our method are works that utilize LLMs as a planner for embodied agents \cite{robot1, robot2, robot3, robot4, robot5, robot6, robot7, unihsi} to generate executable plans in real-world environments. Unlike works focus on robots, we introduce LLMs to manipulate the generation of key poses of motion, enabling fine-granularity control.

\section{Method} \label{sec:method}

\begin{figure*}[htbp]
    \centering
    \includegraphics[width=1.0\textwidth]{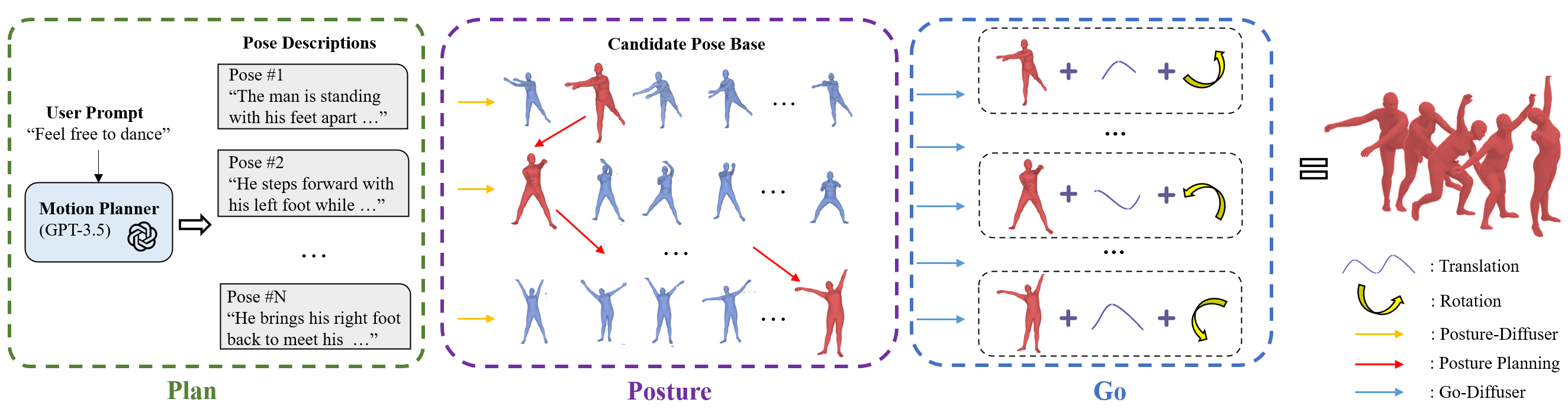}
    \caption{\textbf{Illustration of our framework for open-world text-to-motion generation from language.} Specifically, we employ the large language models as the motion planner to plan pose scripts. Then, our \textit{Posture-Diffuser} module receives discrete pose descriptions and generates corresponding poses to construct a candidate posture base. \textit{Posture Planning} is utilized to select the reasonable pose sequence from the candidate posture base. Finally, the Go-Diffuser module increases the motion frames and infers the translation and rotation.}
    \label{fig:overview}
\end{figure*}

\subsection{Preliminaries}  \label{subsec:preliminaries}

\textbf{Denoising Diffusion Probabilistic Models (DDPMs).} DDPMs, as detailed in~\cite{ddpm1, ddpm2, ddpm3}, involve two Markov chains: a forward chain that diffuses data to noise, and a reverse chain that transforms noise back to data. Formally, the diffusion forward process iteratively adds Gaussian noise to the original signal $x_0$ to generate a sequence of noised samples, \ie, $\{x_t\}^{T}_{t=1}$, which can be formulated as follows:
\begin{equation}
    q(x_t|x_{t-1}) = {\cal{N}}(x_t;{\sqrt{1-\beta_t}}x_{t-1},\beta_tI),
\end{equation}
where $ q(x_t|x_{t-1})$ represents the conditional probability distribution of $x_t$ conditioned by $x_{t-1}$ and $\beta_t \in (0, 1)$ denotes the level of noise at time step $t$. When $T$ is sufficiently large, we can assume that $x_T \sim {\cal N}(0, I)$.

For the conditional synthesis setting, the denoiser is designed to model the distribution $p(x_0|c)$ by sampling a random noise from the prior distribution ${\cal N}(0, \rm{I})$, followed by the reversed diffusion process to convert it back to $x_0$. In our setting, we follow~\cite{ddpm4} and directly predict the original sample, i.e., $\hat{x}_0=f_\theta(x_t, t, c)$ with the simple objective~\cite{ddpm1}:
\begin{equation}
     \underset{\theta}{\text{min}} \ {\cal L}=\mathbb{E}_{x_0, t, c} \ \left[||x_0 - f_\theta(x_t, t, c)||_2^2 \right].
\end{equation}

According to~\cite{ddpm1}, at each time step $t$, we predict the original sample $\hat{x}_0=f_\theta(x_t, t, c)$ and diffuses it back to $x_{t-1}$. The sampling from $p(x_0|c)$ is carried out iteratively, starting from $t = T$, until $x_0$ is obtained. To train our diffusion model $f_\theta$, we employ classifier-free guidance~\cite{classifier} by randomly setting $c = \O$ for 10\% of the samples. This allows $f_\theta$ to learn both the conditioned and the unconditioned distributions, making $f_\theta(x_t, t, \O)$ approximate $p(x_0)$. We use the coefficient $w$ to balance the trade-off between sample diversity and fidelity:
\begin{equation}
    f_\theta^w(x_t, t, c) = f_\theta(x_t, t, \O) + w \cdot (f_\theta(x_t, t, c) - f_\theta(x_t, t, \O)). 
\end{equation}

\subsection{Motion Planner} \label{subsec:motion planner}

As depicted in \cref{fig:overview}, when presented with a user prompt, such as \textit{``Feel free to dance"}, we exploit GPT-3.5~\cite{instructgpt} to create a plan for describing key poses based on the prior knowledge about body parts that are involved in the motion. To ensure that GPT-3.5 generates descriptions for these key poses while maintaining consistency throughout the motion, we provide GPT-3.5 with a user prompt indicating the expected motion and a task description that guarantees the temporal consistency and control over various motion attributes like frames per second (FPS) and the number of frames. Beyond governing the overall motion, we have established five fundamental rules to guide GPT-3.5 in describing key poses: 
(1) Characterize the degree of bending of body parts, \eg, \textit{`left elbow'} using descriptors like \textit{`completely bent', `slightly bent', `straight'.} 
(2) Classify the relative distances between different body parts, 
\eg, two hands, as \textit{`close’}, \textit{`shoulder width apart’}, \textit{`spread’} or \textit{`wide’} apart. 
(3) Describe the relative positions of different body parts, \eg, \textit{`left hip'} and \textit{`left knee'}, using terms like \textit{`behind', `below'} or \textit{`at the right of'}. 
(4) Determine whether a body part is oriented \textit{`vertical'} or \textit{`horizontal'}, \eg, \textit{`right knee'}.
(5) Identify whether a body part is in contact with the ground, such as \textit{`left knee'} and \textit{`right foot'}. Furthermore, we offer GPT-3.5 some reference pose descriptions to guide its generation process. For more details about the prompt design, please refer to the supplementary materials.

\subsection{Posture-Diffuser} \label{subsec:posture diffuser}

In this section, we present our \textit{Posture-Diffuser} module, which aims to generate key poses that align with the localized body part descriptions provided by the \textit{Motion Planner} in \cref{subsec:motion planner}. As shown in \cref{fig:dual-diffusion} (a), we utilize a denoising diffusion model, which is composed of a stack of $N$ identical layers. Each layer has two sub-blocks. The first is a residual block, which incorporates the time embedding generated by passing the sinusoidal time embedding through a two-layer feed-forward network. The second is a cross-modal transformer block, which integrates the conditioning signal, \ie, text, via a standard cross-attention mechanism~\cite{attention}. The intermediate residual pose feature serves as the query vector, while the text embeddings extracted from the frozen DistillBERT~\cite{distilbert} act as the key and value vectors. Furthermore, we randomly mask the text embeddings for classifier-free learning. This module enables us to generate key poses that align with the pose descriptions precisely.

Due to the sampling diversity of DDPMs~\cite{ddpm1, ddpm2, ddpm3, ddpm4}, the \textit{Posture-Diffuser} module can generate multiple plausible poses corresponding to each pose description, we introduce our \textit{Posture Planning} module, which aims to select the most reasonable key poses from the candidate poses. We propose two objectives: (1) minimizing differences between poses in adjacent frames and (2) maximizing the similarity between poses and corresponding descriptions. To achieve the objectives, we design two encoders: a text encoder $\Phi$ composed of a single layer of bi-GRU~\cite{gru}, and a pose encoder $\Theta$ that uses the VPoser encoder~\cite{vposer}. Both the encoders produce the L2-normed embeddings for computing similarity. We employ the Viterbi algorithm~\cite{viterbi} to search for the most reasonable pose path.

Specifically, suppose we have a set of $F$ pose descriptions denoted as $\{t_i\}_{i=1}^{F}$, and for each pose description $t_i$, we have a collection of $L$ generated candidate poses represented as $\{p^{i}_{j}\}_{j=1}^{K}$, serving as pose observations at each frame. The transition probability matrix $A^i$ for the $i$-th ($i > 1$) frame is for the first objective, where the selection of poses for adjacent frames should preferably consider pairs with higher similarity as follows:
\begin{equation}
    A^i_{jk} = \frac{\text{exp}\left({\Theta{(p^{i-1}_{j})}^T\Theta{(p^{i}_{k})}}\right)}{\sum_{l=1}^{L}\text{exp}\left({\Theta{(p^{i-1}_{j})}^T\Theta{(p^{i}_{l}})}\right)}.
\end{equation}
Similarly, the emission probability matrix $E^{i}$ for the $i$-th ($i \geq 1$) frame is to satisfy the second objective, where the selection of the current frame's key pose should preferably consider poses with a higher matching degree to the description as below:
\begin{equation}
    E^{i}_{j} = \frac{\text{exp}\left(\Phi{(t_i)}^T \Theta{(p^{i}_{j})}\right)}{\sum_{l=1}^{L}{\text{exp}\left(\Phi{(t_i)}^T{\Theta{(p^{i}_{l})}}\right)}}.
\end{equation}
The overall objective of the algorithm is to generate a pose path $G = \{{g_i}\}_{i=1}^{F}$ that maximizes the joint probability:
\begin{equation}
    \underset{G}{\text{argmax}} P(G) = \prod_{i=1}^{F} P(g_i|g_{i-1}) = E^{1}_{g_1} \prod_{i=2}^F E^{i}_{g_i}A^{i}_{g_{i-1}g_{i}}.
\end{equation}

\subsection{Go-Diffuser} \label{subsec:go diffuser}

\begin{figure}[tbp]
    \centering
    \includegraphics[width=0.45\textwidth]{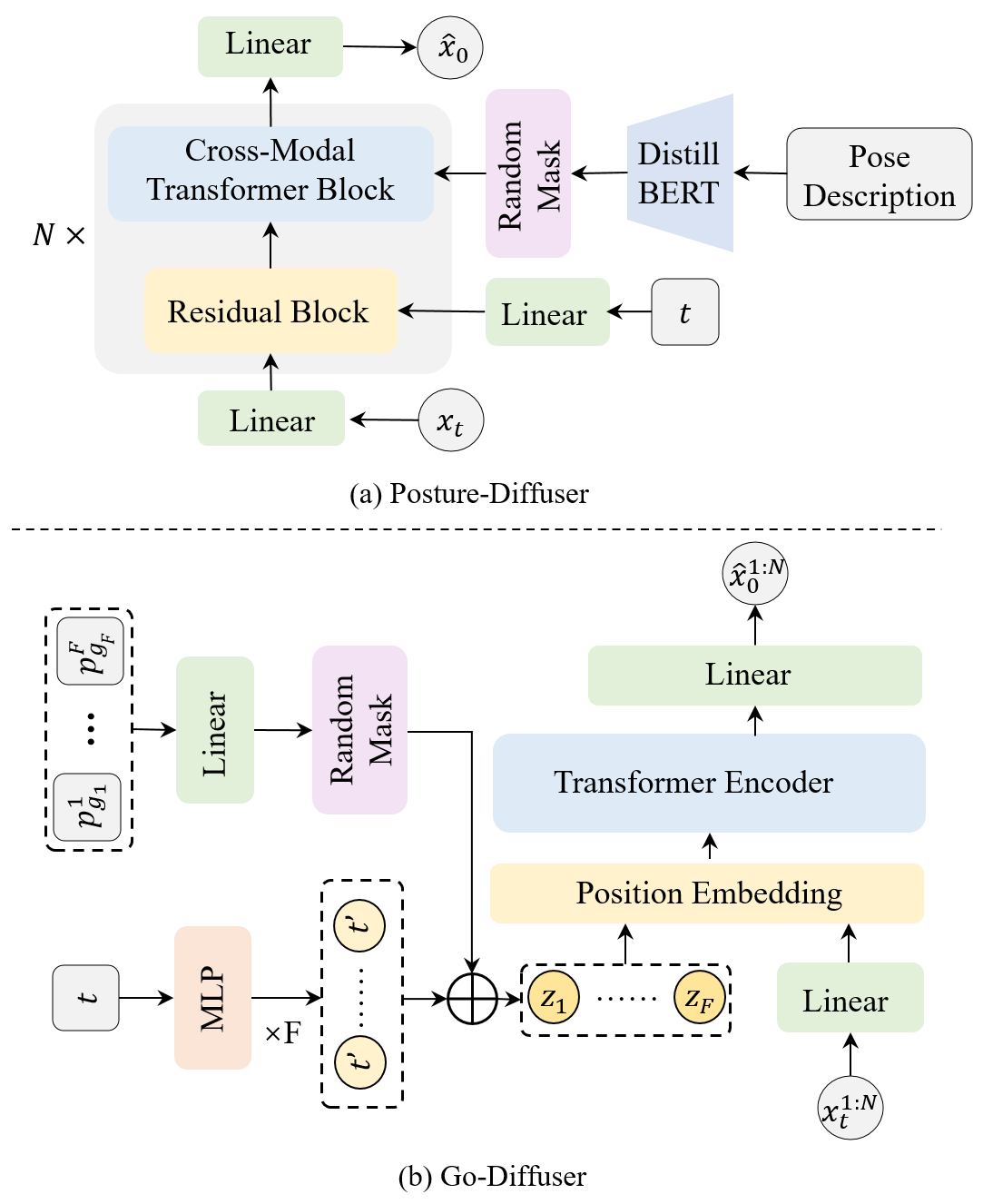}
    \caption{\textbf{Illustration of our Dual-Diffusion model.} 
    (a) \textit{Posture-Diffuser} module is designed to predict the original pose conditioned by the pose description. The model consists of $N$ identical layers, with each layer featuring a residual block for incorporating time step information and a cross-modal transformer block for integrating the condition text.
    (b) \textit{Go-Diffuser} module serves the function of obtaining motion with translation and rotation from discrete key poses without global information. In this module, the key poses obtained from \cref{subsec:posture diffuser} are regarded as independent tokens. We perform attention operations~\cite{attention} between these tokens and noised motion independently, which can significantly improve the perception ability between every condition pose and motion sequence. }
    \label{fig:dual-diffusion}
\end{figure}

To interpolate and predict global information such as translation and rotation for the key poses obtained in \cref{subsec:posture diffuser}, we introduce our \textit{Go-Diffuser} module in this section. Our module is based on a diffusion model, as illustrated in \cref{fig:dual-diffusion} (b). As transformer~\cite{attention} structure has been proved efficient in the field of motion generation~\cite{actor, temos, mdm, tmr}, we adopt it with the transformer encoder architecture in our implementation. The module is fed a noised motion sequence $x^{1:N}_t$ in a time step $t$, as well as $t$ itself and the condition, \ie, key poses ${\{p^i_{g_i}}\}_{i=1}^{F}$. To enhance the modeling of relationships between key poses and better capture global information, we treat them as discrete tokens rather than a unified feature. In practice, the key poses are projected and then randomly masked for classifier-free learning. The noised input $x^{1:N}_t$ is projected and integrates positional information. The transformer encoder output is projected back to the original motion dimension, yielding the predicted motion sequence $\hat{x}^{1:N}_0$. This module enables us to interpolate the key poses smoothly and assign global properties to motions.

\section{Experiments}\label{sec:experiments}

\subsection{Datasets} \label{subsec:datasets}

In our experiments, we utilize the pose data, motion data, and textual descriptions sourced from AMASS~\cite{amass}, PoseScript~\cite{posescript}, Motion-X~\cite{motionx},  and HumanML3D~\cite{humanml3d}. AMASS unifies various optical marker-based mocap datasets, offering over 40 hours of motion data without textual descriptions. PoseScript consists of static 3D human poses extracted from AMASS, together with fine-grained semantic human-written annotated descriptions (PoseScript-H) and automatically generated captions (PoseScript-A). HumanML3D is a widely used motion language dataset that provides captions for motion data sourced from AMASS. Motion-X is a large-scale 3D expressive whole-body motion dataset with detailed language descriptions. We select two subsets from Motion-X for conducting open-world text-to-motion generation experiments. Specifically, we employ sentence transformers~\cite{sentence-bert, sentence-bert-2} to compute the similarity between the text descriptions in IDEA-400~\cite{motionx}, a high-quality motion language subset within Motion-X, and the text descriptions in HumanML3D. We filter out pairs with similarity greater than a specified threshold $\alpha$, \eg, 0.45, yielding a motion language dataset comprising 368 text-motion pairs as our first test dataset. Moreover, we choose the \textit{kungfu} subset of Motion-X as our second test dataset.

\subsection{Open-World Motion Generation} \label{subsec:open-world motion generation}

\begin{table*}[htbp]
\centering
\caption{Comparison of our method with previous methods on the subsets of the IDEA-400~\cite{motionx} dataset, \ie, \textit{ood368} and \textit{kungfu}. We achieve superior performance on R precision, and the MultiModal Dist. MDM~\cite{mdm} is for supervised learning. MotionCLIP~\cite{motionclip}, Codebook+Interpolation~\cite{avatarclip}, Avatarclip~\cite{avatarclip} and OOHMG~\cite{oohmg} are designed for open vocabulary text-to-motion generation.}
\scalebox{1.0}{
\begin{tabular}{l|ccccccc}
\hline 
& \multicolumn{4}{c}{\textbf{Text-motion}} & & \multicolumn{1}{l}{} & \\
\multirow{-2}{*}{} & R@10 $\uparrow$ & R@20 $\uparrow$ & R@30 $\uparrow$ & MedR $\downarrow$ & \multirow{-2}{*}{FID $\downarrow$} & \multicolumn{1}{l}{\multirow{-2}{*}{MultiModal Dist $\downarrow$}} & \multirow{-2}{*}{Smooth $\rightarrow$} \\ \hline \multicolumn{1}{l}{\textit{\textbf{``test on ood368 subset"}}} \\ \hline 
{\color[HTML]{000000} MDM~\cite{mdm}} & {\color[HTML]{000000} 17.81} & {\color[HTML]{000000} 34.06} & {\color[HTML]{000000} 48.75} & {\color[HTML]{000000} 31.20} & {\color[HTML]{000000} 3.500541}          & 2.613644                                                           & {\color[HTML]{000000} 0.000114}          \\ \hline
MotionCLIP~\cite{motionclip}                  & 16.25                        & 35.62                        & 52.81                        & 28.90                        & 2.227522                                 & 2.288905                                                           & 0.000073                                 \\
Codebook+Interpolation~\cite{avatarclip}      & 15.62                        & 31.25                        & 46.56                        & 32.80                        & 4.084785                                 & 2.516041                                                           & 0.000146                                 \\
AvatarCLIP~\cite{avatarclip}                  & 15.31                        & 31.56                        & 47.19                        & 32.60                        & 4.181952                                 & 2.449695                                                           & 0.000146                                 \\ 
OOHMG~\cite{oohmg}     &  15.62                & 34.06            & 48.75                        & 29.80                        & 3.982753                                 & 2.149275                                                           & 0.000758                                 \\ \hline
\textbf{Ours}               & \textbf{20.25}               & \textbf{36.56}               & \textbf{53.14}               & \textbf{26.10}               & {\color[HTML]{000000} \textbf{1.488678}} & \textbf{1.534521}                                                  & {\color[HTML]{000000} 0.001312} \\ \hline

\multicolumn{1}{l}{\textit{\textbf{``test on kungfu subset"}}} \\ \hline
                        
{\color[HTML]{000000} MDM~\cite{mdm}} & {\color[HTML]{000000} 12.50} & {\color[HTML]{000000} 29.69} & {\color[HTML]{000000} 42.19} & {\color[HTML]{000000} 37.50} & {\color[HTML]{000000} 12.060187}   & 3.725436                                                           & {\color[HTML]{000000} 0.000735}        \\ \hline
MotionCLIP~\cite{motionclip}                  & 15.62                       & 29.69                        & 46.88                        & 32.50                        & 17.414746                          & 4.297871                                                           & 0.000123                               \\
Codebook+Interpolation~\cite{avatarclip}      & 10.94                       & 20.31                        & 29.69                        & 37.50                        & 2.521690                           & 2.764137                                                           & 0.000138                               \\
AvatarCLIP~\cite{avatarclip}                  & 15.62                       & 31.25                        & 46.88                        & 32.50                        & 1.966764                           & 2.497678                                                           & 0.000171                               \\ 
OOHMG~\cite{oohmg}     & 14.06                  & 32.81                        & 48.44                        & 32.50                        & 4.904853                                 & 2.471666                                                           & 0.000847                                 \\ \hline
\textbf{Ours}               & \textbf{20.31}              & \textbf{34.38}               & \textbf{50.00}               & \textbf{31.00}               & {\color[HTML]{000000} 4.124218}    & \textbf{2.374380}                                                  & {\color[HTML]{000000} 0.001559}        \\ \hline
\end{tabular}}\label{tab:ood}
\end{table*}

In this section, we first introduce the supervised learning baseline~\cite{mdm, motionclip}, open-world baselines~\cite{motionclip, avatarclip, oohmg}, and the evaluation metrics \cite{actor, humanml3d}. Then we discuss the comparative experimental results with these baselines.

\noindent \textbf{MDM Baseline}: MDM~\cite{mdm} employs a supervised learning approach utilizing the diffusion model. However, its performance often degenerates when applied to a new setting, \ie, open-vocabulary motion generation. We trained it on the SMPL-H~\cite{smpl, smplx, mano} version data of AMASS with HumanML3D annotation. 

\noindent \textbf{MotionCLIP Baseline}: MotionCLIP~\cite{motionclip} is a supervised open-vocabulary method which trained on AMAS data with BABEL~\cite{babel} annotation. We use the pre-trained model provided by the authors and test the model's performance in the open-vocabulary setting. 

\noindent \textbf{Codebook+Interpolation Baseline}: In the pose generation stage, we utilize VPoserCodebook~\cite{smplx} as the pose generator and select the most similar pose in the pose generation stage. For the motion generation stage, we just use the interpolation method to generate the motion.

\noindent \textbf{AvatarCLIP Baseline}: AvatarCLIP~\cite{avatarclip} is an optimizer-based method. It also includes a text-to-pose stage via matching and uses the matched poses to search the most related motion in the latent space of a motion VAE~\cite{vae} trained on the AMASS dataset.

\noindent \textbf{Evaluation Metrics} is adopted from~\cite{actor, humanml3d}, which includes R precision, Frechet Inception Distance(FID), and MultiModal Distance. For quantitative evaluation, a motion feature extractor and a text feature extractor are trained using contrastive loss to produce geometrically close feature vectors for matched text-motion pairs. For more details about the above metrics as well as the design of the text and motion feature extractor, please refer to the supplementary materials. Consider R precision: for each generated motion, its ground-truth text description and randomly selected mismatched descriptions from the test dataset form a description pool, followed by calculating and ranking the Euclidean distances between the motion feature and the text feature of each description in the pool. Meanwhile, MultiModal distance is computed as the average Euclidean distance between the motion feature of each generated motion and the text feature of its corresponding description in the test dataset.

\begin{figure*}[htbp]
    \centering
    \includegraphics[width=1\textwidth]{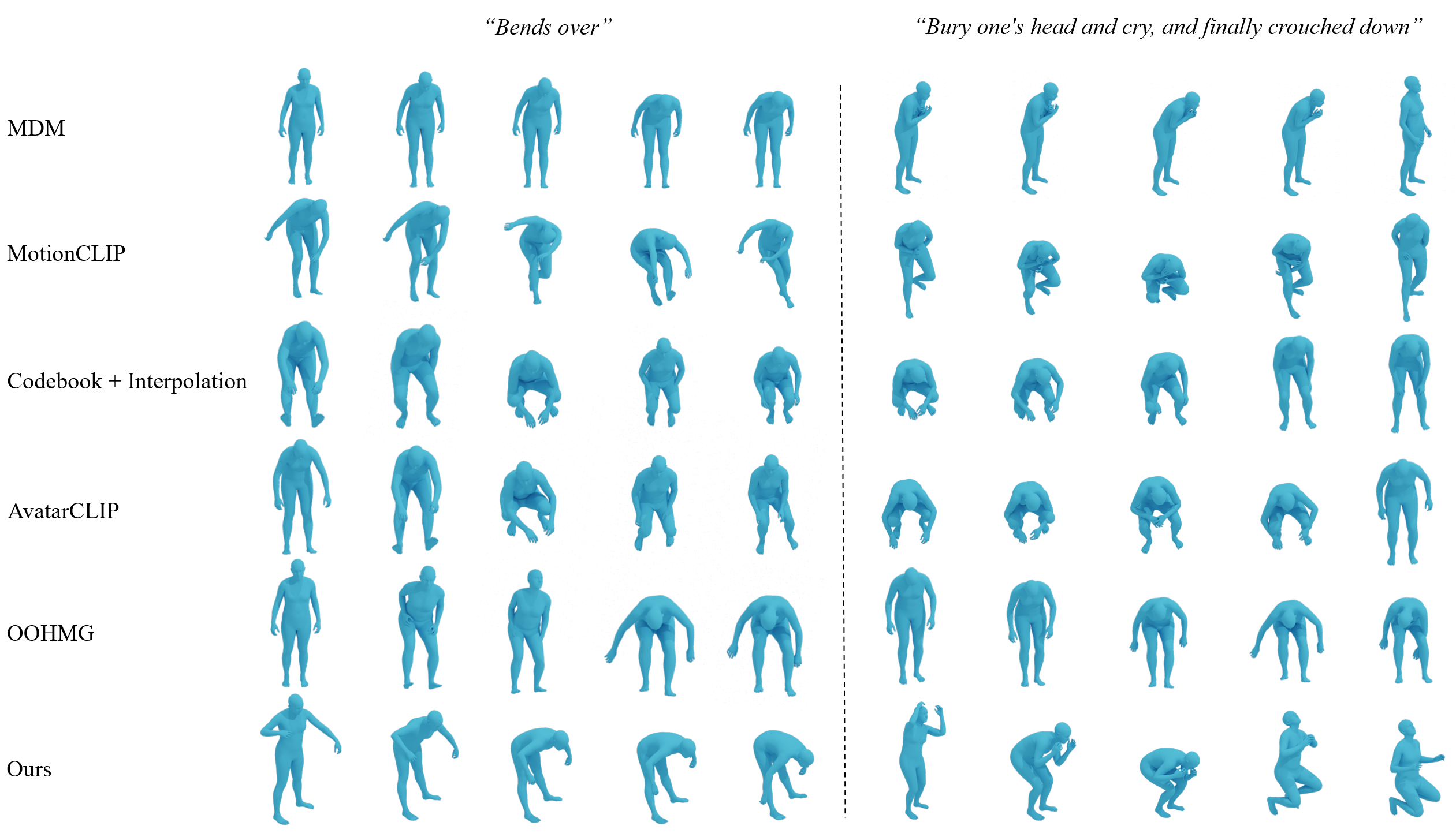}
    \caption{Comparation of our methods with previous text-to-motion generation methods.}
    \label{fig:t2m}
\end{figure*}

As shown in \cref{fig:t2m}, for the motion description ``bends over", the supervised-based method MDM~\cite{mdm} generated the reasonable motion sequence, which shows the process from upright to prone. While OOHMG~/cite{oohmg} involves more leg movement, which is not accurate for the description. Other methods utilizing CLIP~\cite{clip}, such as MotionCLIP~\cite{motionclip}, Codebook+Interpolation~\cite{avatarclip}, and AvatarCLIP~\cite{avatarclip} mistake the key pose sequence, their spine first curves to a certain degree, and then the degree of bending decreases. Due to the lack of temporal priors in CLIP, the generated pose sequence in the first stage does not follow the right order. Moreover, for the motion description \textit{``bury one's head and cry, and finally crouched down"}, methods such as MDM~\cite{mdm} based on the supervised learning paradigm often fail in similar cases and cannot generate un-seen motion. Due to the gap between motion description and image description, matching text and motion via the language space of CLIP is not effective. They struggle to deal with detailed and precise motion descriptions. In \cref{tab:ood}, quantitative metrics demonstrate the superiority of our model over other methods in terms of semantic consistency and motion rationality.

\begin{table*}[]
\centering
\caption{Comparison of our method with baseline methods on AMASS\cite{amass} dataset. We achieve state-of-the-art performance on Average Positional Error and Average Variance Error. \textit{Root joint, global traj. and mean local} metrics represent the performance of translation and rotation. \textit{Mean global} represents the performance of body joints and global translation. }
\begin{tabular}{l|cccc|cccc}
\hline
\multirow{2}{*}{Methods} & \multicolumn{4}{c|}{Average Positional Error $\downarrow$}        & \multicolumn{4}{c}{Average Variance Error $\downarrow$}           \\ 
                         & root joint & global traj. & mean local & mean global & root joint & global traj. & mean local & mean global \\ \hline
Regression               &      5.878673      &      5.53344        &     0.642252       &    5.919954         &    35.387340        &      35.386562        &    0.147606        &    35.483219         \\ \hline
Baseline\cite{actor}                   & 0.384152   & 0.373394     & 0.183978   & 0.469322    & 0.114308   & 0.113845     & 0.015207   & 0.126049    \\ \hline
Ours                     & \textbf{0.365327}   & \textbf{0.354685}     & \textbf{0.128763}   & \textbf{0.418265}    & \textbf{0.111131}   & \textbf{0.110855}     & \textbf{0.008708}   & \textbf{0.118334}    \\ \hline
\end{tabular} \label{table:p2m}
\end{table*}

\subsection{Ablation Study}

\begin{figure*}[htbp]
    \centering
    \includegraphics[width=1\textwidth]{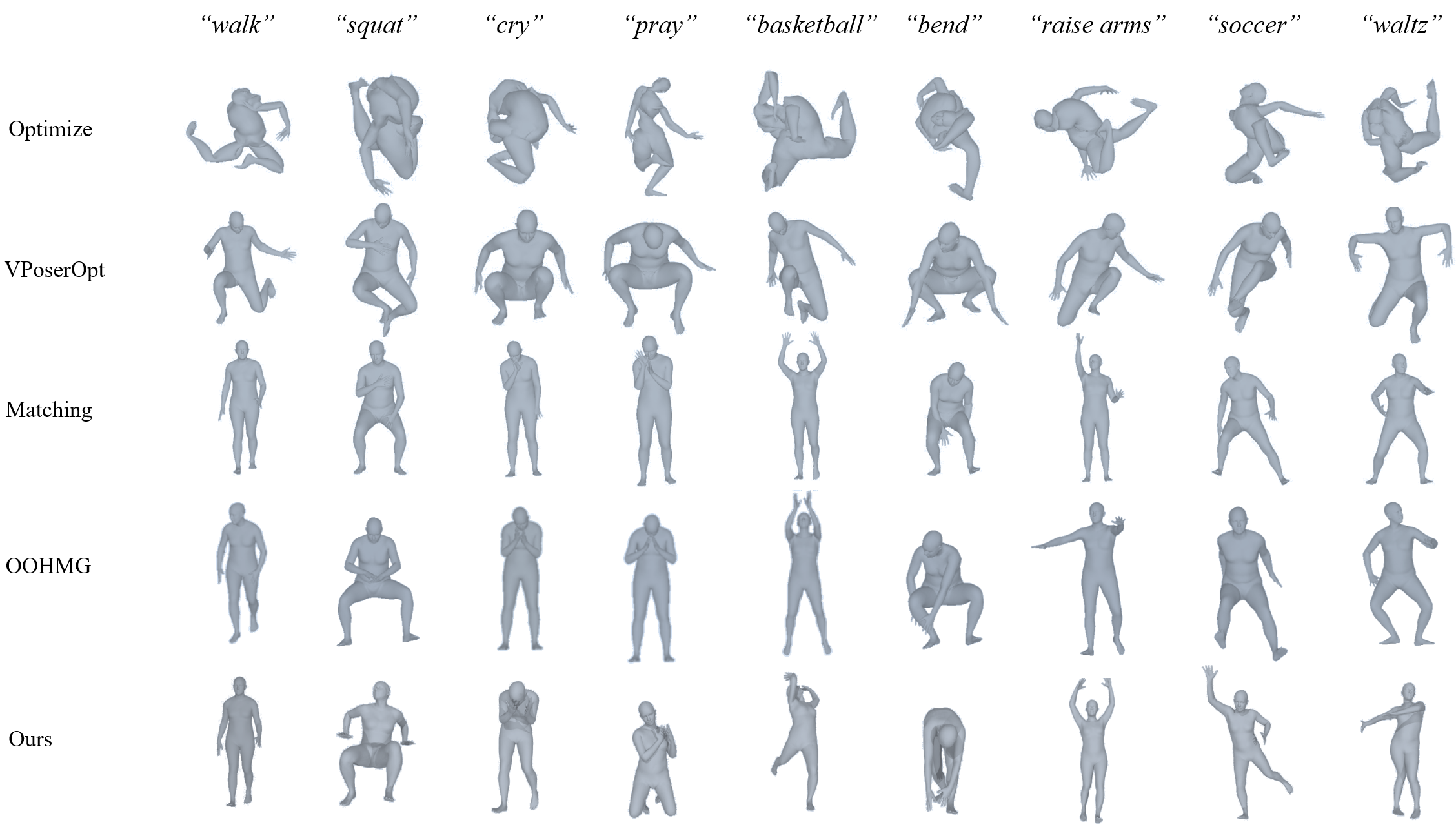}
    \caption{Comparison of our method with previous text-to-pose generation methods.}
    \label{fig:t2p}
\end{figure*}

\textbf{Posture-Diffuser} In comparison to the zero-shot open-vocabulary text-to-pose generation methods~\cite{avatarclip, oohmg}, we first utilize LLMs to translate the pose description into localized body part description, which is fed to our pose generator in \cref{subsec:posture diffuser} to generate pose precisely. As shown in \cref{fig:t2p}, the Matching method demonstrates superior pose generation results compared to Optimize and VPoserOptimize, which suggests that directly using CLIP for matching is more effective than optimization through the complex pipeline. However, ``Matching" fails to generate more precise poses for diverse texts, exhibiting a limitation in preserving textual information in the generated poses. For instance, in cases like \textit{``cry"} and \textit{``pray"}, ``Matching" generates identical poses for texts with distinct meanings. When generating poses that require more precise control over body parts, such as \textit{``dance the waltz"} or \textit{``kick soccer"}, both OOHMG and “Matching” fail to achieve satisfactory results. In contrast, by employing LLMs to precisely describe the expected pose, we achieve accurate control over pose generation, enabling more effective open-world text-based pose synthesis.

\begin{figure*}[htbp]
    \centering
    \includegraphics[width=1.0\textwidth]{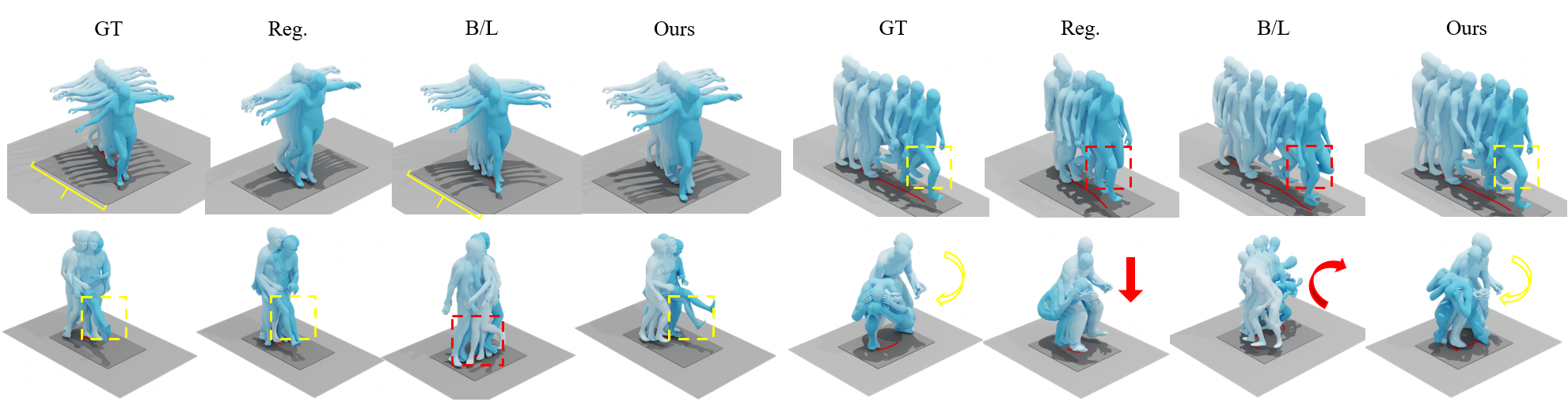}
    \caption{Comparison of different methods. Yellow color represents details that need attention, and red color represents inaccuracies.}
    \label{fig:p2m}
\end{figure*}

\noindent \textbf{Go-Diffuser.} While this is the first effort at predicting spatial information of motion in a zero-shot local pose-driven manner to our best knowledge, we have developed appropriate baseline methods to evaluate the translation and rotation perception power. Based on our observation, the translation and rotation of the people could be estimated by analyzing the variations in body parts between adjacent poses. We treat the process of estimating global information from the key poses as a reconstruction task. Thus, we designed a simple MLP-based network to achieve this first. The transformer structure has been proven correct in the motion generation filed~\cite{temos, actor}, thus we design the baseline method by extracting pose sequence features as the condition to inject into the diffusion model. As shown in \cref{fig:p2m}, from the four images in the top left corner, it can be observed that a simple MLP network is capable of predicting motion translation information to some extent. As indicated in the bottom-left image, extracting pose sequences as features using existing motion encoders may overlook the internal relationships within the pose sequences, thereby leading to confusing details. In the top right image, our method exhibits better fidelity in capturing fine details such as knee flexion. Moreover, the four images in the bottom right corner reveal that when dealing with similar adjacent poses, our model demonstrates a finer-grained perceptual capability, thus imparting appropriate motion trends to digital avatars. As shown in \cref{table:p2m}, our method achieved state-of-the-art results both on the average positional error and the average variance error of global trajectory, rotation and local pose joints.

\section{Conclusion}
\label{sec:conclusion}
In this paper, we introduce PRO-Motion, a model designed to tackle open-world text-to-motion generation tasks. It consists of three modules: motion planner, posture-diffuser, and go-diffuser. The motion planner instructs the large language models to generate a sequence of scripts describing the key postures in the target motion. The posture-diffuser transforms a script into a posture, paving the way for open-world generation. Finally, the go-diffuser, estimates whole-body translations and rotations for all postures, resulting in diverse and realistic motions. Experimental results have shown the superiority of our method compared to other counterparts.
{\small
\normalem
\bibliographystyle{ieeenat_fullname}
\bibliography{11_references}
}

\ifarxiv \clearpage \appendix 
\definecolor{codegreen}{rgb}{0,0.6,0}
\definecolor{codegray}{rgb}{0.5,0.5,0.5}
\definecolor{codepurple}{rgb}{0.58,0,0.82}
\definecolor{backcolour}{rgb}{0.95,0.95,0.92}

\lstdefinestyle{mystyle}{
  backgroundcolor=\color{backcolour}, commentstyle=\color{codegreen},
  keywordstyle=\color{magenta},
  numberstyle=\tiny\color{codegray},
  stringstyle=\color{codepurple},
  basicstyle=\ttfamily\footnotesize,
  breakatwhitespace=false,         
  breaklines=true,                 
  captionpos=b,                    
  keepspaces=true,                 
  numbers=left,                    
  numbersep=5pt,                  
  showspaces=false,                
  showstringspaces=false,
  showtabs=false,                  
  tabsize=2
}

\lstset{style=mystyle}
\noindent{\textbf{\fontsize{14.0pt}{\baselineskip}\selectfont Appendix}}\vspace{1.5ex}

\noindent In this supplementary material, we provide additional details and experiments not included in the main paper due to limitations in space.
\begin{itemize}
    \item \cref{supp:pose}: Qualitative results of precise control for motion generation in our formulation.
    \item \cref{supp:motion}: Details of the motion representations. 
    \item \cref{supp:prompt}: Details of the prompt engineering and the entire system message template.
    \item \cref{supp:model}: Details of our dual-diffusion models.
    \item \cref{supp:metric}: Details of the evaluation metrics.
\end{itemize}
\noindent Note: \textcolor{blue}{Blue} characters denote the main paper's reference.
\section{Precise Pose Control} \label{supp:pose}
As shown in \cref{fig:t2p-supp}, our \textit{Posture-Diffuser} module can accurately generate poses from edited pose descriptions, which becomes particularly beneficial when users are not content with the key pose descriptions produced by the \textit{Motion Planner}. In such cases, users have the option to manually edit the description of specific body parts to gain precise control over the generated poses. For example, in the Control $\#$1 of Pose Description $\#$1, we replace the description of ``their left arm is behind the right with their right elbow bent at 90 degrees." with ``the left hand is beside the right arm and the right hand is to the left of the left hand". It's obvious the position of the hands changes correctly. As well, when we delete the description of ``The left foot is stretched backwards and is behind the right foot" in the Control $\#1$ of Pose Description $\#$2, the position of the left foot changes correctly. Furthermore, we could find that although the scripts are simple and limited to a small space, they are expressive to cover all possible postures due to their compositional nature.

\begin{figure*}[tbp]
    \centering
    \includegraphics[width=\textwidth]{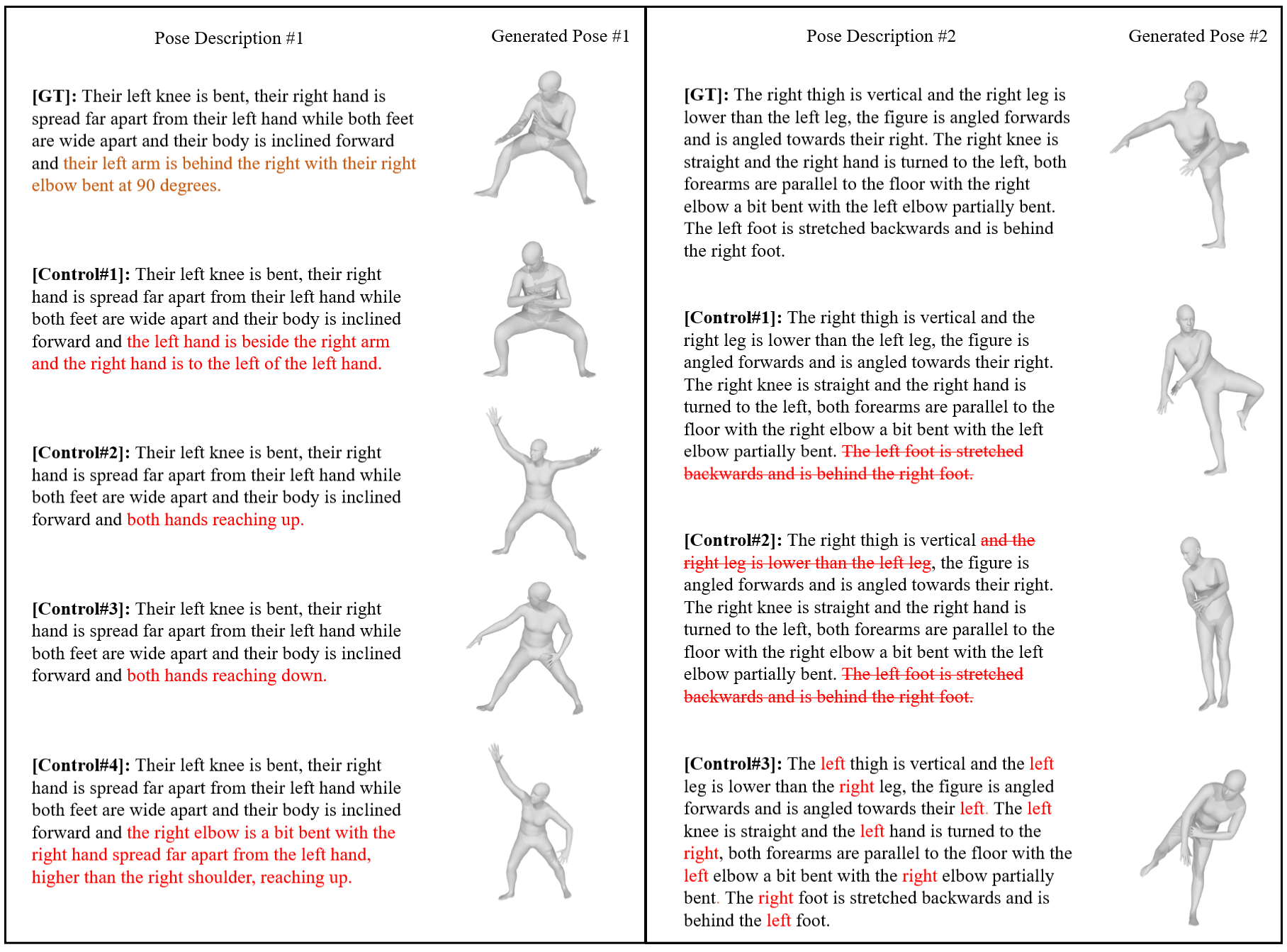}
    \caption{Examples of Precise Pose Control. In the first example \#1, we control the position of hands, with the original and modified hand descriptions represented in {\color{brown} brown} and {\color{red} red} respectively. In the second example \#2, ``{\color{red}{\sout{The}}}" represents deleting the description. To determine whether the description of a body part is the reason for the proper positioning of that body part, we removed the corresponding descriptions to see if the body joints would change as a result of this operation. We test the model's understanding of the left and right sides of the body.}
    \label{fig:t2p-supp}
\end{figure*}

\section{Motion Representations} \label{supp:motion}
Skinned Multi-Person Linear model (SMPL)\cite{smpl}\cite{mano} is a skinned vertex-based model that accurately represents various body shapes in natural human poses. It's widely utilized in the human motion generation field. So we adopted the SMPL-based human model as well. SMPL begins with an artist-created mesh with N = 6890 vertices and K = 23 joints. The mesh has the same topology for men and women, spatially varying resolution, a clean quad structure, a segmentation into parts, initial blend weights, and a skeletal rig. 
We follow the SMPL-based representation of TEMOS~\cite{temos} and construct the feature vector for SMPL data.
\begin{itemize}
    \item \textbf{Translation.} It consists of two parts. The first part is the velocity of the root joint in the global coordinate system. The second part is the position of the root joint for the Z axis. The dimension of the translation is 3.
    \item \textbf{Root Orientation.} It contains one rotation, and we utilize the 6D continuous representation~\cite{rot} to store it. So the dimension of the root orientation is 1 * 6 = 6.
    \item \textbf{Pose Body.} The motion data is from the SMPL-H~\cite{smpl, mano} version of AMASS. Because we focus on the movement of the human body, we removed all rotations on the hands, resulting in 21 rotations). The same as root orientation, we utilize the 6D continuous representation~\cite{rot}. So the dimension of the pose body is 21 * 6 = 126.
\end{itemize}

So the final dimension of the feature is $3 + 1 * 6 + 21 * 6 = 135$. In our experiments, we remove all the motion sequences that are less than 64 frames, and the motion sequences longer than 64 frames are processed as 64 frames. The data sample will be in $ \mathbb{R}^{64 * 135} $.

\begin{lstlisting}[language=Python, caption=motion data]
# read motion 
motion_parms = {
    'trans': motion[:, :3], # controls the global body position
    'root_orient': motion[:, 3 : 3 + 6], # controls the global root orientation
    'pose_body': motion[:, 9 : 9 + 21 * 6], # controls the body
}
\end{lstlisting}

When we perform the translation and rotation prediction, we only know the local attributes of the adjacent poses, but not their global location information. So we must normalize the pose position. Following ~\cite{temos}, the translation of neighboring poses is subtracted and represented by the instantaneous velocity as the translation attribute of the current frame. 

\begin{lstlisting}[language=Python, caption=translation normalization]
# extract the root gravity axis
# for smpl it is the last coordinate
root = trans[..., 2]
trajectory = trans[..., [0, 1]]

# Comoute the difference of trajectory (for X and Y axis)
vel_trajectory = torch.diff(trajectory, dim=-2)
# 0 for the first one => keep the dimentionality
vel_trajectory = torch.cat((0 * vel_trajectory[..., [0], :], vel_trajectory), dim=-2)
\end{lstlisting}


\section{Prompt Design} \label{supp:prompt}

\cref{fig:prompt} illustrates the complete prompt used by our \textit{Motion Planner} in \cref{subsec:motion planner}. We first define the overall objective and task requirements and then establish five fundamental rules of body parts including bending degree, relative distance, relative position, orientation, and ground contact. Additionally, we specify the body parts to which each rule applies. Through this rule-based approach, we can guide the LLM to generate precise key pose descriptions, achieving fine-grained control over poses. Next, we define the output format of the LLM and provide examples of pose descriptions for the LLM to reference. These examples are sourced from Posescript~\cite{posescript}. Finally, a user prompt of motion description is provided to instruct the LLM to generate the key pose descriptions. By harnessing the powerful LLM, the user prompts are no longer confined to specific formats but are open-world, allowing for expressions like \textit{``Jump on one foot"} and \textit{``Experiencing a profound sense of joy"}.

\begin{figure*}[tbp]
    \centering
    \includegraphics[width=0.95\textwidth]{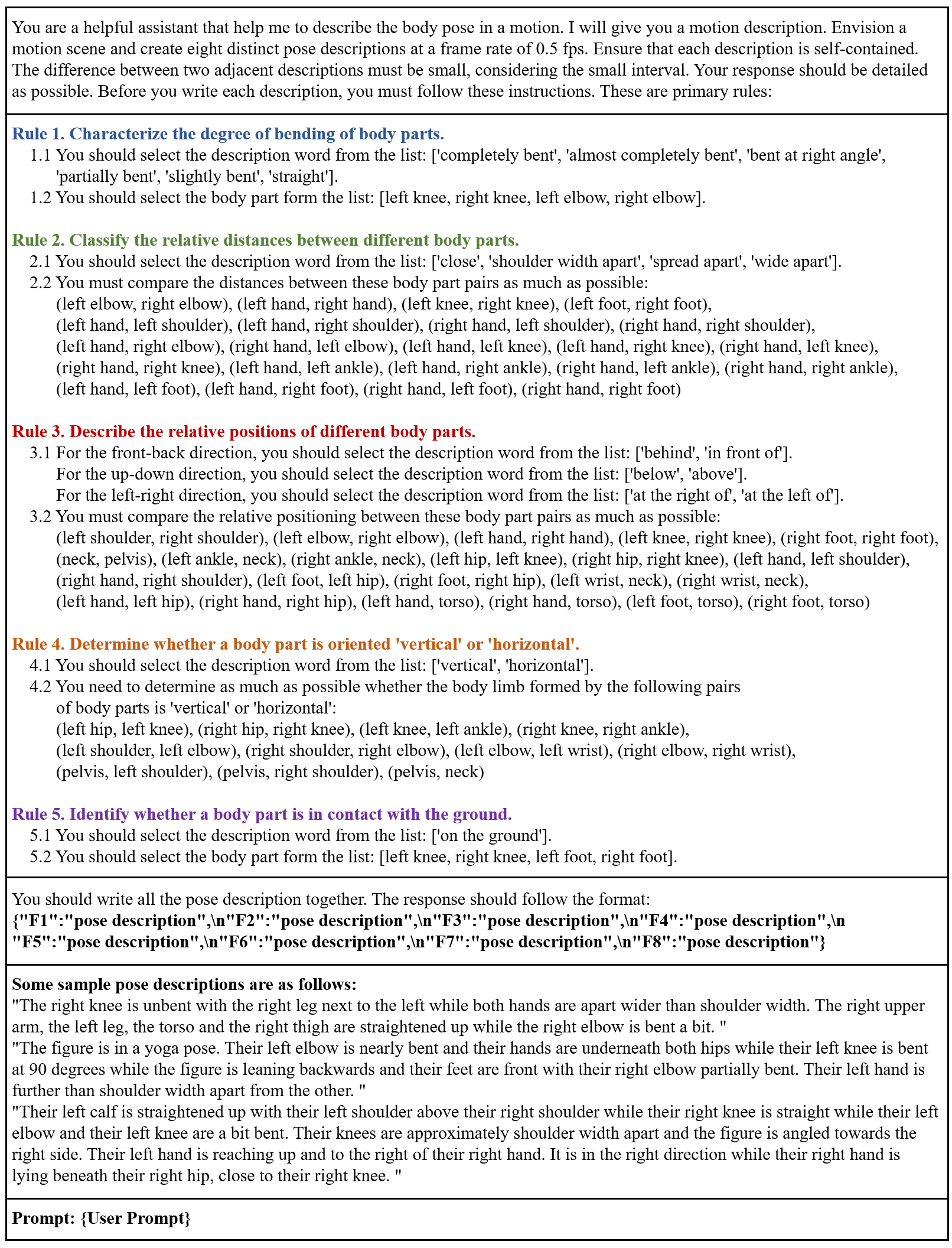}
    \caption{The complete prompt for our \textit{Motion Planner} in \cref{subsec:motion planner}.}
    \label{fig:prompt}
\end{figure*}

\section{Detailed Implementations} \label{supp:model}
\textbf{Posture-diffuser}. We employ an AdamW~\cite{adamw} optimizer for 1000 epochs with a batch size of 512 and a learning rate of 1e-4. The number of layers is $N =3$ and the latent dim is set to 512. Following~\cite{ddpm1}, we use the \textit{linear} schedule, where $\beta_{start} = 0.00085$ and $\beta_{end} = 0.012$. Our model is trained using poses with automatically generated captions (PoseScript-A~\cite{posescript}) and adopts the last checkpoint as our final model. For the two encoders used in \textit{Posture Planning} module, we use the pretrained text-to-pose retrieval model from~\cite{posescript}, which is trained on PoseScript-A using the Batch-Based Classification loss~\cite{bbc}.

\noindent \textbf{Go-diffuser.} We employ an AdamW~\cite{adamw} optimizer with a batch size of 64 and a learning rate of 1e-4. The latent dim is 512, the number of transformer layers is 8, and the number of heads is 4. We utilize GELU~\cite{gelus} as the activation function and the dropout is 0.1. The mask probability of the model's condition is 0.1.  We use the cosine schedule and the number of diffusion steps is 100.

\section{Evaluation details}\label{supp:metric}
We will detail our evaluation metrics on the Experiments part in this section. 
In open-world text-to-motion experiments, we employ sentence transformers~\cite{sentence-bert, sentence-bert-2} to compute the similarity between the text descriptions in IDEA-400~\cite{motionx}, a high-quality motion language subset within Motion-X, and the text descriptions in HumanML3D. We filter out pairs with similarity greater than a specified threshold $\alpha$, \eg, 0.45, yielding a motion language dataset comprising 368 text-motion pairs as our first test dataset. 

\begin{lstlisting}[language=Python, caption=similarity filter]
# calculate the similarity of motion descriptions and filter out pairs with similarity greater than a specified threshold
import torch
import torch.nn as nn
import torch.nn.functional as F
from transformers import AutoTokenizer, AutoModel, util

def mean_pooling(model_output, attention_mask):
    token_embeddings = model_output[0]  # First element of model_output contains all token embeddings
    input_mask_expanded = attention_mask.unsqueeze(-1).expand(token_embeddings.size()).float()
    return torch.sum(token_embeddings * input_mask_expanded, 1) / torch.clamp(input_mask_expanded.sum(1), min=1e-9)
    
# calculate the features of descriptions
class TextToSen(nn.Module):
    def __init__(self, device):
        super().__init__()
        self.device = device
        self.tokenizer = AutoTokenizer.from_pretrained('sentence-transformers/all-mpnet-base-v2')
        self.model = AutoModel.from_pretrained('sentence-transformers/all-mpnet-base-v2').eval().to(self.device)

    def forward(self, sentences):
        encoded_input = self.tokenizer(sentences, padding=True, truncation=True, return_tensors='pt').to(self.device)

        # Compute token embeddings
        with torch.no_grad():
            model_output = self.model(**encoded_input)

        # Perform pooling
        sentence_embeddings = mean_pooling(model_output, encoded_input['attention_mask'])

        # Normalize embeddings
        sentence_embeddings = F.normalize(sentence_embeddings, p=2, dim=1)

        return sentence_embeddings
        
# calculate the similarity
similarity_score = util.pytorch_cos_sim(embeddings1, embeddings2)

if similarity_score > threshold:
    False

\end{lstlisting}

\textbf{R-Precision} is determined by ranking the Euclidean distances between the motion and text embeddings, given one motion sequence and K text descriptions (1 ground-truth and $K-1$ randomly selected mismatched descriptions).

\textbf{Frechet Inception Distance} (FID) measures the distributional difference between the generated and real motion by applying FID~\cite{gans} to the extracted motion features derived from that text.

\textbf{Multimodal Distance} (MM-Dist) is calculated as the average Euclidean distance between each text feature and the corresponding generated motion feature.

In go-diffuser experiments, we utilize Average Position Error (APE) and Average Variance Error (AVE) to evaluate our methods~\cite{temos}. We report the listed four metrics in \cref{table:p2m}.

\noindent \textbf{Root joint errors.} Take the 3 coordinates of the root joint.

\noindent \textbf{Global trajectory errors.} Tak3 only the X and Y coordinates of the root joint. It is the red trajectory on the ground in the visualizations of ~\cref{fig:p2m}.

\noindent \textbf{Mean local errors.} Average the joint errors in the body's local coordinate system,  

\noindent \textbf{Mean global errors.} Average the joint errors in the global coordinate system.

As in JL2P~\cite{language2pose}, Ghosh et al.~\cite{compositional_animations} and TEMOS~\cite{temos}, the APE for a specific joint j is determined by computing the mean of the L2 distances between the generated and ground truth joint positions across the frames (F) and samples (N):
\begin{equation}
    APE[j]=\frac{1}{NF} \sum_{n\in N}^{} \sum_{f\in F}^{} \left \| H_{f}\left [ j \right ] -\hat{H}_{f}\left [ j \right ]    \right \|_{2} 
\end{equation}
As introduced in Ghosh et al~\cite{compositional_animations} and TEMOS~\cite{temos}, the Average Variance Error (AVE), quantifies the distinction in variations. This metric is defined as the mean of the L2 distances between the generated and ground truth variances for the joint j.
\begin{equation}
    AVE[j]=\frac{1}{N} \sum_{n\in N}^{}  \left \| \delta \left [ j \right ] - \hat{\delta}  \left [ j \right ]    \right \|_{2}     
\end{equation}
where,
\begin{equation}
     \delta \left [ j \right ] = \frac{1}{F-1} \sum_{f \in F}^{} \left ( H_{f}\left [ j \right ] - \tilde{H}_{f}\left [ j \right ]   \right ) ^{2} \in R^{3} 
\end{equation}
denotes the variance of the joint j. 


 \fi

\end{document}